\documentclass{article}
\usepackage{spconf,amsmath,graphicx}

\title{Supervised level-wise pretraining for Recurrent Neural Network initialization in multi-class classification}
\name{Dino Ienco$^{\star }$ \qquad Roberto Interdonato$^{\dagger}$ \qquad Raffaele Gaetano$^{\dagger}$}

			\address{$^{\star}$ IRSTEA, UMR TETIS, LIRMM, Univ. of Montpellier, Montpellier \\
			  $^{\dagger}$ CIRAD, UMR TETIS, Montpellier}
\begin{document}
%
\maketitle
\begin{abstract}
Recurrent Neural Networks (RNNs) can be seriously impacted by the initial parameters assignment, which may result in poor generalization performances on new unseen data.
With the objective to tackle this crucial issue, in the context of RNN based classification, we propose a new supervised layer-wise pretraining strategy to initialize network parameters. The proposed approach 
leverages a data-aware strategy that sets up a taxonomy of classification problems automatically derived by the model behavior. To the best of our knowledge, despite the great interest in RNN-based classification, 
this is the first data-aware strategy dealing with the initialization of such models. 
The proposed strategy has been tested on four benchmarks coming from two different domains, i.e.,  Speech Recognition and Remote Sensing. Results underline the significance of our approach and point out that data-aware strategies positively support the initialization of Recurrent Neural Network based classification models.

\end{abstract}

\section{Introduction}
\label{sec:intro}

Deep Learning based approaches are commonly adopted techniques to cope with a broad variety of signal analysis scenarios: Convolutional Neural Networks (CNN) have demonstrated their effectiveness in a plethora of different image analysis tasks~\cite{LiuWLZLA17} while Recurrent Neural Networks (RNN) have exhibited impressive results in one dimensional signal processing tasks such as Natural language processing and understanding, Speech Recognition classification and Remote sensing analysis~\cite{KhalilJBJZA19,MouGZ17,IencoGDM17}.
Despite the fact that new neural network architectures are constantly proposed~\cite{HeZRS16,HuangLMW17,CaiZH19}, the success of such approaches is a result of many advances in basic elements such as activation functions and initialization strategies. During the training process, neural networks are sensitive to the initial weights~\cite{PengKRP18}. In particular, RNNs can suffer from ``vanishing gradient'' or instability if they are not correctly initialized~\cite{GlorotB10}. In the context of RNN, few initialization approaches were proposed~\cite{ChoromanskiDB18} which are generally task or data agnostic. Other strategies initialize network weights by exploiting pretrained models learned on big datasets~\cite{GeY17,DevlinCLT19} but 
such pretrained networks are not available in general cases. As shown in~\cite{PengKRP18} for supervised classification tasks, neural network initialization is a critical issue that can prejudice good generalization performances on new unseen data.

With the aim to deal with RNN initialization, in this paper we propose a data/task-aware supervised pretraining strategy to initialize network parameters. 
Differently from existing approaches~\cite{ChoromanskiDB18}, where RNN models are initialized in a data agnostic setting, here we propose to leverage the basic RNN model behavior on the task at hand to set up a taxonomy of nested classification problems, ranked by decreasing order of complexity. Following such hierarchy of classification tasks, weights are learnt and transferred through the different levels in a top-down fashion by means of sequential fine-tuning. Finally, the weights derived by such sequence of successive discrimination problems will be used to initialize the RNN based model dealing with the original classification task.
Our approach can be seen as a self-pretrain/fine-tuning pipeline, which is adaptively built based on the specific task and dataset under analysis. 
To the best of our knowledge, the only work that tackles the issue of weight initialization considering a supervised data-aware setting is the one in~\cite{PengKRP18}. Here, the authors propose a strategy to support the initialization of a multi-layer perceptron based model using a supervised pretraining involving the discrimination between real vs shuffled data, with no adaptation with respect to the specific task (data-aware but not task-aware).

To assess the performance and the behavior of our proposal, we conduct experiments on four benchmarks coming from two different domains, comparing our results with those obtained using several baseline and competing approaches.
\section{TAXO: Supervised level-wise pretraining for RNN-based classification}
\label{sec:method}

In this section we introduce our supervised level-wise pretraining procedure to initialize the parameters weight of a generic Recurrent Neural Network dealing with multi-class classification tasks. We denote with $X = \{ X_i \}_{i=1}^n$ the set of multi-variate time series data. We underline that $X$ can contain time-series with different lengths.
We can divide the set of examples $X$ in three disjoint sets: training set $X^{train}$, validation set $X^{valid}$ and test set $X^{test}$ where $X^{train} \cap X^{valid} = \emptyset$, $X^{train} \cap X^{test} = \emptyset$ and $X^{valid} \cap X^{test} = \emptyset$. Associated to $X^{train}$ and $X^{valid}$ we also have a class information: $Y^{train}$ and $Y^{valid}$, respectively. $Y^{train}$ and $Y^{valid}$ are defined on the same set of classes: $Y_i^{train} \in C$ and $Y_i^{valid} \in C$ where $C=\{C_j\}_{j=1}^{|C|}$ is the set of classes and $|C|$ is the number of different classes.
The supervised level-wise pretraining process has three steps driven by the task at hand in conjunction with the behavior of the model on the associated data:

\noindent
\textbf{Step 1.} Given a training and a validation set ($X^{train}$ and $X^{valid}$) with the corresponding class set ($Y^{train}$ and $Y^{valid}$), the Recurrent Neural Network model is trained on ($X^{train}$, $Y^{train}$) while ($X^{valid}$, $Y^{valid}$) is used to perform an unbiased evaluation of the model while tuning the model parameters. The model that achieves the best performances on $X^{valid}$ across a fixed number of training epochs is retained. Here, the classification task is performed on the original set of classes $C$. The RNN model generated at this step is named $M_{0}$.

\noindent
\textbf{Step 2.}  Given the retained model $M_{0}$, we compute the confusion matrix between the prediction of $M_{0}$ on $X^{valid}$ and $Y^{valid}$. Successively, for each class $C_j$, we compute the entropy associated to the prediction of that class: \small $$ Entropy(C_j) = - \sum_{k=1}^{|C|} p_{jk} \times \log{p_{jk}}$$ \normalsize where $p_{jk}$ is the probability of the class $j$ to be misclassified with the class $k$. Such probability is obtained from the confusion matrix after it has been normalised by row to obtain a probability distribution with row-sums equal to 1. The entropy quantifies the uncertainty of the model $M_{0}$ on each class, i.e., a class that is misclassified with many other ones is associated with a bigger uncertainty value than a class that is always misclassified with just another class. Given that higher entropy values correspond to higher classifier uncertainty, the classes $C_j$ are then ranked in decreasing order w.r.t. their uncertainty, producing the ranking $R$.

\noindent
\textbf{Step 3.}  We exploit the ranking $R$ to produce a taxonomy of different classification problems on which the RNN will be pretrained for the multi-class classification tasks defined on the set of classes $C$. 
Given a depth parameter $d$, we proceed level-wise: we pick the first class of the ranking $R$, $C_{R(1)}$, and we build a binary classification task in which the training and validation examples ($X^{train}$ and $X^{valid}$) associated to the class $C_{R(1)}$ have the label $1$ while all the other examples have the label $0$. We hence train the model (from scratch at this level), and generate the current model $M_{1}$. Then, we select the class in the second position of the ranking, $C_{R(2)}$, and add another class to the classification problem (3 classes at level 2). We fine-tune the model $M_{1}$ on the new three-class classification task obtaining the model $M_{2}$. The procedure continues level-wise until the level $d$ is reached. 

Generally speaking, the model $M_{t}$ is trained (fine-tuned for $t>1$) on a multi-class problem involving $t+1$ classes, where the first $t$ classes are those in the set $\bigcup_{j=1}^{t} C_{R(j)}$ with the higher values of entropy and the last class groups together all the others original classes ($C \setminus \{ \bigcup_{j=1}^{t} C_{R(j)} \}$). At the end of this process, the weights of the model $M_{d}$ are employed to initialize the final RNN model that will tackle the original multi-class classification task defined on the set of classes $C$. Note that we deliberately choose to leave the maximum depth $d$ as a free parameter, as a mean to control the computational burden of the method, but ideally the whole depth inferred by the total number of classes ($d = |C|-1$) might be explored.

In practice, each classification stage is achieved by connecting the Recurrent Neural Network unit (LSTM or GRU) directly with an output layer with as many neurons as the number of involved classes. The decision is obtained via a SoftMax layer. Considering the main step of our approach, in which the different models ($M_{1}, ..., M_{d}$) are learnt, we only modify the output layer at each different level with a dense output layer with a number of neurons equal to the corresponding number of classes. The parameter weights that are transferred from one level of the taxonomy to another are the sole parameters of the RNN unit.

The rationale behind our strategy is that it supports the parameter learning exploration to focus on classes that are difficult to predict,  modifying in a second time the previous parameters to integrate the discrimination of ``easier'' classes in the model decision. In other words, it forces the learning trajectory to pass through the weaknesses of the dataset, in order to avoid biasing the final model towards the recognition of the strongest (separable, representative) classes.

\section{Experimental evaluation}
\label{sec:expe}
In this section we evaluate our supervised pretraining strategy on four real-world datasets coming from two different domains: Speech Recognition and Remote Sensing (Satellite Image Time Series) classification. In both domains, RNNs are a common state of the art strategy to perform supervised classification~\cite{KhalilJBJZA19,IencoGDM17}. For the former domain Long-Short Term Memory (LSTM) is the RNN model that is usually employed~\cite{KhalilJBJZA19} while Gated Recurrent Unit have demonstrated their ability to well fit the characteristics of remote sensing data~\cite{IencoGDM17,Interdonato19}. Considering both LSTM and GRU, we equipped the RNN models with an attention mechanism~\cite{BritzGL17}, a widely used strategy to allow RNN models to focus on the most informative portions of the input sequences. We name the RNN models equipped with our supervised level-wise pretraining strategy $LSTM_{taxo}$ and $GRU_{taxo}$.
As baseline competitors, we take into account the following approaches: 

i) An RNN approach without any supervised pretraining, named $LSTM$ ($GRU$ resp.) for Speech Recognition (resp. Remote Sensing) task. 

ii) An RNN approach with a supervised pretraining inspired by~\cite{PengKRP18}. First, a copy of the dataset is created by independently shuffling, for each sample, the order of its elements in the sequence to break sequential dependencies. Then a binary classification task is set up to discriminate between examples coming from the original dataset and examples coming from the shuffled dataset. The network is firstly trained to solve the binary task. Finally, the weights of the network that have dealt with the binary task are employed as initialization of the RNN network dealing with the original multi-class task. We name such a competitor $LSTM_{sh}$ ($GRU_{sh}$ resp.).

iii) A variant of the proposed supervised pretraining approach in which the loss function, at the different levels of pretraining, is weighted using class cardinalities in order to cope with class unbalance (larger weights to classes with fewer samples). We name such competitor $LSTM_{taxo}^{W}$ ($GRU_{taxo}^{W}$ resp.).

To assess the performances of the different methods we consider two standard metrics: Accuracy and F-Measure. The metrics are computed performing a train, validation and test evaluation considering 50\%, 20\% and 30\%, respectively, of the original data. For each dataset and method, we repeat the procedure 5 times and we report average and standard deviation for each metric.
For all the experiments, the number of pretraining levels for our approach is fixed to 3. We use Adam as optimizer to learn the parameters weights with a learning rate of 5~$\times$~10$^{-4}$ and a batch size of 32.

\subsection{Speech Recognition Tasks}
We consider two well known speech recognition tasks: Japanese Vowel (\textit{JapVowel}) and Arabic Digits (\textit{ArabDigits}) (cf. Table~\ref{tab:dataSpeech}). The former sets up a classification task to distinguish among nine male speakers by their utterances of two Japanese vowels while the latter involves 88 different speakers (44 males and 44 females) pronouncing digits in Arabic language. For each benchmark, Frequency Cepstral Coefficients (MFCCs) are firstly extracted and, successively, multi-variate time series are constructed for each sample in the dataset. Considering the setting for this evaluation, all the Recurrent Neural Network (LSTM) approaches have 256 hidden units and they are trained for 250 epochs.

\begin{table}[!htb]
    \centering
    \scriptsize
    \begin{tabular}{|c|c|c|c|c|} \hline
        \textbf{Dataset} & \textbf{\# Samples} & \textbf{\# Dims} & \textbf{Min/Max/Avg Length} & \textbf{\# Classes} \\ \hline
         JapVowel & 640 & 12 & 7/29/15 & 9 \\  \hline
         ArabDigits & 8\,800 & 13 & 4/93/39 & 10 \\ \hline
    \end{tabular}
    \caption{Speech Recognition datasets characteristics \label{tab:dataSpeech}}
    \label{tab:my_label}
\end{table}

\begin{table}[!ht]
    \centering
    \scriptsize
    \begin{tabular}{|c||c|c||c|c|} \hline
        & \multicolumn{2}{c|}{\textbf{JapVowel}} & \multicolumn{2}{|c|}{\textbf{ArabDigits}} \\ \hline
        & F-Measure & Accuracy & F-Measure & Accuracy \\ \hline
        LSTM & 94.53 $\pm$ 1.82 & 94.61 $\pm$ 1.74 & 99.42 $\pm$ 0.16 & 99.42 $\pm$ 0.16 \\ \hline
        LSTM$_{Sh}$ & 94.82 $\pm$ 1.67 & 94.84 $\pm$ 1.67 & \textbf{99.69 $\pm$ 0.23} & \textbf{99.69 $\pm$ 0.23} \\ \hline
        LSTM$_{taxo}^{W}$ & \textbf{95.53 $\pm$ 1.61} & \textbf{95.53 $\pm$ 1.60} & 99.58 $\pm$ 0.09 & 99.58 $\pm$ 0.09\\ \hline
        LSTM$_{taxo}$ & 94.91 $\pm$ 2.80 & 95.05 $\pm$ 2.60 & 99.60 $\pm$ 0.10 & 99.60 $\pm$ 0.10 \\ \hline
    \end{tabular}
    \caption{F-Measure and Accuracy, average and std. deviation, on the Speech Recognition tasks considering the different competing methods.\label{tab:resSpeech}}
\end{table}

Table~\ref{tab:resSpeech} reports on the Accuracy and F-Measure of the different competing methods on the two Speech Recognition datasets. We can observe that all the supervised pretraining methods (LSTM$_{Sh}$, LSTM$_{taxo}^{W}$ and LSTM$_{taxo}$ ) generally achieve higher performances than the baseline (LSTM). On the \textit{JapVowel} benchmark, the two variants of our proposal obtain the best results, with  LSTM$_{taxo}^{W}$ being the best performing method. This can be explained by the fact that, on such small sized dataset (640 samples), random sampling produces train, validation and test subsets with slightly different class distributions, making the loss weighting strategy more appropriate for generalization. 
On the \textit{ArabDigits} speech recognition task, the best performances are obtained by the LSTM$_{Sh}$ approach with a very limited gain in terms of both F-Measure and Accuracy w.r.t. the competitors.
Nevertheless, although the baseline strategy already achieves high discrimination performances on this dataset, we can see that all our supervised pretraining approaches reach comparable scores, while slightly reducing the classification uncertainty compared to LSTM$_{Sh}$.

\subsection{Remote Sensing Data Analysis}
We consider two Satellite Image Time Series (SITS) land cover classification datasets: \textit{Gard} and \textit{Reunion} Island (cf. Table~\ref{tab:dataRS}). Both datasets are preprocessed as done in~\cite{IencoIGARSS19}: starting from a suitable object layer, the spectral information is aggregated at segment level for each available timestamp, hence generating the multi-dimensional time series. The first dataset (\textit{Gard})
~\cite{Interdonato19}, concerns a zone in the South of France, while the second one (\textit{Reunion})~\cite{IencoIGARSS19}, focuses on Reunion Island, a French overseas department located in the Indian Ocean. 
For both datasets, source imagery is acquired via the ESA Sentinel-2 (S2) mission. The first dataset \textit{Gard} uses a larger spectral information (10 bands from S2 at 10 and 20~m resolution, all resampled at 10~m) plus six radiometric indices~\cite{Interdonato19}, while the second dataset \textit{Reunion} only includes four radiometric bands at 10m of resolution plus the Normalized Differential Vegetation Index~\cite{IencoIGARSS19} for a total of 5 channels for each timestamps.
 As regards the neural network, for this evaluation all the Recurrent Neural Network (GRU) approaches have 512 hidden units and they are trained for 1\,000 epochs.

\begin{table}[!ht]
    \centering
    \small
    \begin{tabular}{|c|c|c|c|c|c|} \hline
         \textbf{Dataset} & \textbf{\# Samples} & \textbf{\# Dims} & \textbf{Length} & \textbf{\# Classes} \\ \hline
         Gard & 1\,673 & 16 & 37 & 7 \\ \hline
         Reunion & 7\,462 & 5 & 34 & 13 \\ \hline
    \end{tabular}
    \caption{Satellite Image Time Series datasets characteristics. \label{tab:dataRS}}
\end{table}

\begin{table}[!ht]
    \centering
    \scriptsize
    \begin{tabular}{|c||c|c||c|c|} \hline
        & \multicolumn{2}{c|}{\textbf{Gard}} & \multicolumn{2}{|c|}{\textbf{Reunion}} \\ \hline
        & F-Measure & Accuracy & F-Measure & Accuracy \\ \hline
        GRU & 85.59 $\pm$ 1.35 & 85.74 $\pm$ 1.36 & 68.34 $\pm$ 0.52 & 70.46 $\pm$ 0.27\\ \hline
        GRU$_{Sh}$ & 84.00 $\pm$ 1.10 & 84.00 $\pm$ 1.05 & 66.45 $\pm$ 2.08 & 68.66 $\pm$ 1.52 \\ \hline
        GRU$_{taxo}^{W}$ & 89.64 $\pm$ 1.64 & 89.74 $\pm$ 1.52 & 73.03 $\pm$ 2.81 & 74.33 $\pm$ 2.41\\ \hline
        GRU$_{taxo}$ & \textbf{90.84 $\pm$ 1.49} & \textbf{90.89 $\pm$ 1.48} & \textbf{76.77 $\pm$ 4.41} & \textbf{77.67 $\pm$ 3.84} \\ \hline
    \end{tabular}
    \caption{F-Measure and Accuracy, average and std. deviation, on the Satellite Image Time Series datasets considering the different competing methods.}
    \label{tab:res_RS}
\end{table}

Results of the evaluation are reported in Table~\ref{tab:res_RS}.
It can be noted how in this case the proposed methods (GRU$_{taxo}$ and GRU$_{taxo}^{W}$) significantly outperform the baseline (GRU). Differently from the previous experiments, we can note that the competitor based on timestamps shuffle (GRU$_{Sh}$) degrades the performances w.r.t. the baseline approach. This is not surprising due to the strong temporal dependencies that characterize such kind of data~\cite{Interdonato19}. Here, the supervised level-wise pretraining clearly ameliorates the performances of the RNN approach and GRU$_{taxo}$ demonstrated to be more effective that its variant involving the weighting loss mechanism (i.e., GRU$_{taxo}^{W}$). 
Note also that, differently to what happened on \textit{JapVowel}, the SITS benchmarks are characterized by a strong class unbalance, with very close distributions on both training and test sets, which makes the unweighted loss strategy more effective. 
 
\subsection{Inspection on the Reunion dataset: Automatic vs Knowledge-based pretraining}
As extra knowledge associated to the \textit{Reunion} dataset, remote sensing domain experts were able to provide a domain hierarchy that structures the land cover classes as a taxonomy. Due to the availability of such information, we derive another competing approach that, following the spirit of the strategy we have introduced in Section~\ref{sec:method}, leverages the domain hierarchy (DH) to perform level-wise supervised pretraining. We name such competitor GRU$_{DH}$.
Table~\ref{tab:DH} summarizes the quantitative results of this new competitor compared to the baseline approach as well as GRU$_{taxo}$. We can observe that, GRU$_{taxo}$ still outperforms the knowledge-based competitor while, we can point out that the knowledge-based (domain hierarchy) competitor provides better performances than the baseline method trained from scratch.

\begin{table}[!ht]
    \centering
    \footnotesize
    \begin{tabular}{|c||c|c|} \hline
         & \multicolumn{2}{|c|}{\textbf{Reunion}} \\ \hline
        & F-Measure & Accuracy \\ \hline
        GRU &  68.34 $\pm$ 0.52 & 70.46 $\pm$ 0.27\\ \hline
        GRU$_{DH}$  & 73.88 $\pm$ 2.40 & 75.08 $\pm$ 1.94 \\ \hline
        GRU$_{taxo}$  & \textbf{76.77 $\pm$ 4.41} & \textbf{77.67 $\pm$ 3.84} \\ \hline
    \end{tabular}
    \caption{F-Measure and Accuracy, average and std. deviation, on the \textit{Reunion} dataset considering a standard GRU, a layer-wise pretraining mechanism exploiting domain hierarchy (GRU$_{DH}$) and the proposed approach (GRU$_{taxo}$). \label{tab:DH} }
    
\end{table}

\begin{figure}[!hbt]
\centering
\includegraphics[width=0.95\columnwidth]{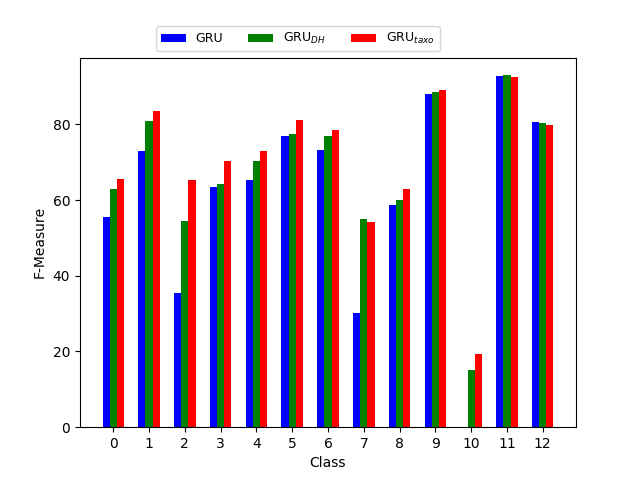}
\caption{\small Per-Class F-measure of the GRU, GRU$_{taxo}$ and GRU$_{DH}$ approaches.
Class ID, Class Name and \# Samples: (0, {\em Crop cultivationd}, 380), (1, {\em Sugar cane}, 496), (2, {\em Orchards}, 299), (3, {\em Forest plantations}, 67), (4, {\em Meadow}, 257 ), (5, {\em Forest}, 292 ), (6, {\em Sh. savannah}, 371 ), (7, {\em Herb. savannah}, 78 ), (8, {\em Rocks}, 107 ), (9, {\em Urb. areas}, 125 ), (10, {\em GreenHouse crops}, 50 ), (11, {\em Water}, 96 ) and (12, {\em Shadow}, 38 ). \normalsize \label{fig:perClassFM}}
\end{figure}

As additional evaluation, we also inspect the per-class F-Measure results of the three methods involved in this experiment: GRU,  GRU$_{DH}$ and GRU$_{taxo}$. Figure~\ref{fig:perClassFM} reports on such results. We can note that, considering almost all the classes, GRU$_{taxo}$ allows to ameliorate the per-class F-Measure performances. We investigate the correlation between class improvements and the most confused classes selected by our layer-wise pretraining strategy, and we observed that such classes benefit from a clear improvement. For instance, inspecting the behavior of our method, we note that classes 0 ({\em Crop cultivation}), 2 ({\em Orchards}) and 7 ({\em Herb. savannah}) are frequently involved in the layer-wise process (they are commonly associated to high value of entropy) and, on such classes, GRU$_{taxo}$ clearly exhibits strong behavior compared to the other competing approaches. 

\vspace{-8mm}
\section{Conclusion}
In this paper we have presented a new supervised level-wise pretraining strategy to support RNN-based classification.  To the best of our knowledge, no previous work was proposed to deal with data/task-aware RNN initialization for classification tasks.
Our strategy has been assessed on benchmarks coming from two different domains,  i.e.,  Speech Recognition and Remote Sensing. On both domains, our proposal has achieved better or comparable performances w.r.t. the competing methods. In addition, we have inspected the performances of our data-aware approach versus a knowledge-based solution in the context of SITS data. While the latter still outperforms the baseline approach, the former achieves even better performances.  Future research will be devoted to transfer the proposed strategy to other neural models, i.e. CNN,  as well as to the compression of the level-wise taxonomy to avoid the choice of the depth parameter $d$.

\bibliographystyle{IEEEbib}
\bibliography{refs}

\end{document}